\documentclass[11pt]{article}
\usepackage{ACL2023}
\usepackage{times}
\usepackage{latexsym}

\usepackage[T1]{fontenc}

\usepackage[utf8]{inputenc}
\usepackage{graphicx}
\usepackage{booktabs}
\usepackage{multirow}
\usepackage{multicol}
\usepackage{color}
\usepackage{latexsym}
\usepackage{xspace}
\usepackage{microtype}
\usepackage{amsmath,amsfonts,amssymb,amsthm, bm}
\usepackage{inconsolata}

\title {TACR: A Table-alignment-based Cell-selection and Reasoning Model for Hybrid Question-Answering}

\author{Jian Wu$^{1}$\footnotemark[1] , Yicheng Xu$^{2}$\footnotemark[1] , Yan Gao$^{3}$, Jian-Guang Lou$^{3}$ \\
\bf Börje F. Karlsson$^{4}$, Manabu Okumura$^{1}$ \\
$^{1}$Tokyo Institute of Technology \quad
$^{2}$Nanyang Technological University \\
$^{3}$Microsoft Research Asia \quad
$^{4}$Beijing Academy of Artificial Intelligence \\
\tt wu.j.as@m.titech.ac.jp, yxu040@e.ntu.edu.sg, \\ 
\tt \{yan.gao, jlou\}@microsoft.com, borje@baai.ac.cn, oku@pi.titech.ac.jp
}

\begin{document}
\maketitle

\renewcommand{\thefootnote}{\fnsymbol{footnote}}
\footnotetext[1]{indicates equal contribution.}
\renewcommand{\thefootnote}{\arabic{footnote}}

\begin{abstract}
Hybrid Question-Answering (HQA), which targets reasoning over tables and passages linked from table cells, has witnessed significant research in recent years. A common challenge in HQA and other passage-table QA datasets is that it is generally unrealistic to iterate over all table rows, columns, and linked passages to retrieve evidence. Such a challenge made it difficult for previous studies to show their reasoning ability in retrieving answers. To bridge this gap, we propose a novel \textbf{T}able-\textbf{a}lignment-based \textbf{C}ell-selection and \textbf{R}easoning model (TACR) for hybrid text and table QA, evaluated on the HybridQA and WikiTableQuestions datasets. In evidence retrieval, we design a table-question-alignment enhanced cell-selection method to retrieve fine-grained evidence. In answer reasoning, we incorporate a QA module that treats the row containing selected cells as context. Experimental results over the HybridQA and WikiTableQuestions (WTQ) datasets show that TACR achieves state-of-the-art results on cell selection and outperforms fine-grained evidence retrieval baselines on HybridQA, while achieving competitive performance on WTQ. We also conducted a detailed analysis to demonstrate that being able to align questions to tables in the cell-selection stage can result in important gains from experiments of over 90\% table row and column selection accuracy, meanwhile also improving output explainability. \footnote{ Code is publicly available at https://github.com/WuJian1995/QAP}
\end{abstract}

\section{Introduction}

\begin{figure*}[ht]
\centering
\includegraphics[width=1.0\textwidth]{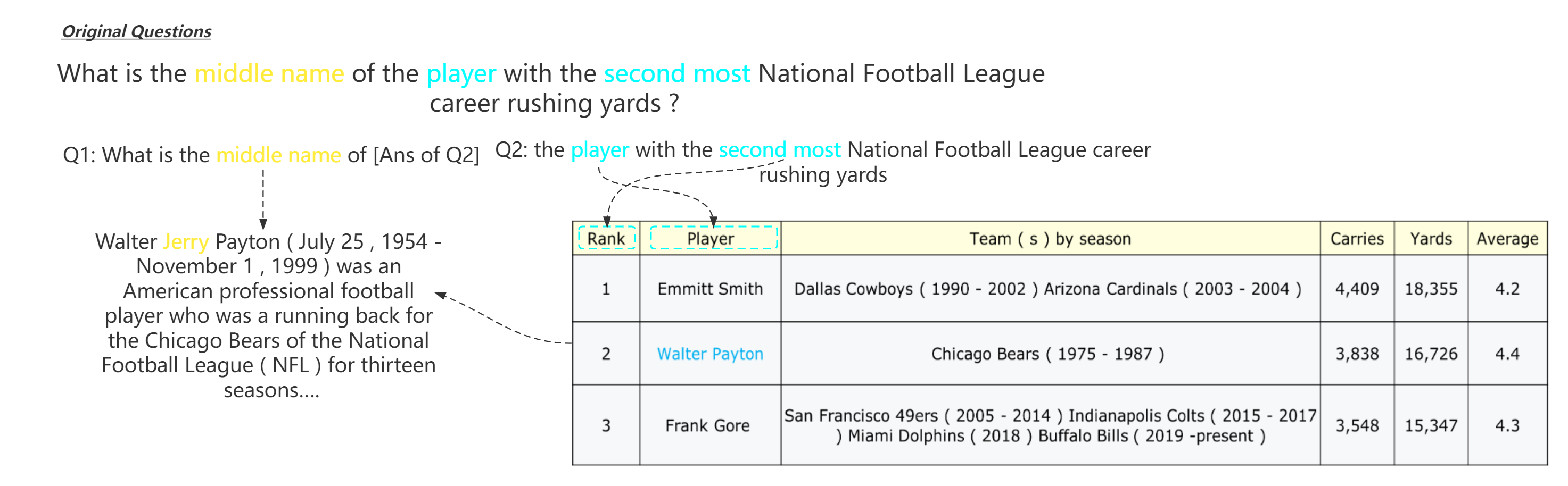}
\caption{\textcolor{black}{Example from the HybridQA dataset. The top sentence is the original question, and words in different colors show different parts of questions required for reasoning in different modalities. the two headers in blue-dashed boxes are column names aligned with the given question. TACR first uses a method based on table-question-alignment to align the original question with table columns to help obtain golden table cells and then retrieve the final answer based on linked passages.}}
\centering
\label{Figure Example}
\end{figure*}

Text-based question-answering datasets derive answers based on reasoning over given passages~\cite{rajpurkar2016squad, chen-etal-2017-reading, Joshi2017TriviaQAAL, yang-etal-2018-hotpotqa}, while table-based QA datasets collect tables from sources such as WikiTables~\cite{pasupat-liang-2015-compositional, zhong2017seq2sql, chen2019tabfact}. However, datasets combining textual passages and tables, like HybridQA~\cite{chen2020hybridqa}, OTT-QA~\cite{Chen2020OpenQA}, and TAT-QA~\cite{zhu2021tat} are more realistic benchmarks. As the answer to a given question may come from either table cells or linked passages, current hybrid QA models usually consist of two components, a retriever to learn evidence and a reasoner to leverage the evidence to derive answers. Such models retrieve evidence from different granularities, coarse-grained (e.g., row or column) or fine-grained (e.g., cell), and directly use a span-based reading comprehension model to reason the answer.  

\citet{Kumar2021MultiInstanceTF}, for example, chooses coarse-grained regions as evidence, e.g., a table row. \citet{chen2020hybridqa} and \citet{Eisenschlos2021MATEMA}, however, focus on fine-grained units, table cells and linked passages. To preserve the advantages and eliminate the disadvantages of different-granularity evidence, \citet{Sun2021EndtoEndMR} propose MuGER,$^2$ which performs multi-granularity evidence retrieval and answer reasoning. 

\citet{Wang2022MuGER2ME} conducts extensive experiments to prove that a coarse-grained retriever contributes less than a fine-grained retriever. Moreover, fine-grained methods, although giving an exact position of candidate cells, fail to illustrate why the selected cells are chosen, while our method is based on row and column selection probabilities. We thus further extend the fine-grained method by aligning questions with tables, letting our approach know which parts of questions are accounted for by which modalities. Intuitively, multi-hop questions in the text-table QA task usually contain two pieces of information from different modalities, tables and passages. Moreover, tables and passages are connected with evidence contained in tabular data. Our method implicitly decomposes the questions for different modalities to locate evidence and improve cell-selection accuracy. 

As illustrated in Figure \ref{Figure Example}, an example from the HybridQA dataset shows how humans work on multi-hop and multi-modal QA tasks. The original question \emph{"What is the middle name of the player with the second most National Football League career rushing yards ?"} can be divided into two parts, \emph{"What is the middle name of"} and \emph{"the player with the second most National Football League career rushing yards?"} for passages and tables, respectively. Such sub-questions are connected with the evidence of a cell ( \emph{"Walter Payton"}). For humans, we first locate who was the player in the second rank, which requires information from two columns: \emph{"Rank"} and \emph{"Player"}. After locating the cell, we can finally determine Walter Payton's middle name from the passage. 
Such reasoning process inspired us to develop TACR, a \textbf{T}able-\textbf{a}lignment-based \textbf{C}ell-selection and \textbf{R}easoning model, which incorporates a fine-grained evidence-retrieval module that utilizes table-question-alignment to learn which parts of the question are used for retrieving evidence from different modalities and reasoning towards answers. 

To explicitly and correctly show the reasoning process in the text-table QA task, in the evidence retrieval stage, TACR first selects the golden cells and avoids redundant information in multi-granularity evidence that would lower the performance of the answer-reasoning module. The table-cell-selection module of TACR is designed to navigate the fine-grained evidence for the reader by fusing well-learned information from the table-question-alignment module. Compared with current fine-grained retrievers, the table-question-alignment module of TACR can help our model learn which parts of questions are used for reasoning in which modality, and which parts of tables contain candidate cells. Together with the alignment module, TACR preserves both high golden cell-selection accuracy and shows competitive performance on the HybridQA and WikiTableQuestions (WTQ) datasets, while providing improved explainability. 

Our contributions are as follows: (1) TACR is the first model able to explicitly show its reasoning process in the passage-table QA task; (2) We jointly train the cell-selection and table-question alignment modules to improve golden cell selection performance and preserve the QA reader’s performance; and (3) We conduct extensive experiments on the HybridQA and WTQ datasets to demonstrate the effectiveness of TACR. 

\section{Related Work}
\subsection{Table Question Answering}
Table QA has gained much attention, as shown by benchmark datasets such as WikiTableQuestions~\cite{Pasupat2015CompositionalSP}, WikiSQL~\cite{Zhong2018Seq2SQLGS}, SPIDER~\cite{Yu2018SpiderAL}, and TABFACT~\cite{chen2019tabfact}. However, these datasets mainly focus on reasoning on tables and ignore important knowledge stored in the textual corpus. Consequently, QA covering both tabular and textual knowledge has gained increasing interest. \citet{chen2020hybridqa} pioneered a passage-table QA benchmark, HybridQA, with Wikipedia tables linked to relevant free-form text passages (e.g., Wikipedia entity-definition pages). The OTT-QA~\cite{Chen2020OpenQA} benchmark extended HybridQA to the open domain setting, where a system needs to retrieve a relevant set of tables and passages first before trying to answer questions. Moreover, the links from the table and passage are not provided explicitly. 

\subsection{Table-Question Alignment}
There are several table-question-alignment methods. Schema-linking-based methods, such as RAT-SQL~\cite{Wang2019RATSQLRS}, introduce a relation-aware transformer encoder to improve the joint encoding of a question and schema. \citet{Liu2022SemanticET} propose a similarity learning-based question-schema-alignment method to obtain a semantic schema-linking graph and observed how the pre-trained language model (PLM) embeddings for the schema items are affected. \citet{Zhao2022TablebasedFV} use the same words that appear in both the natural language statement and the table as weak supervised key points and design an interaction network to explore the correlation between the representations of the natural language statements and tables. 

\subsection{Hybrid QA}
Studies on hybrid QA usually retrieve different granularities of evidence from heterogeneous data to retrieve the final answer. Hybrider, proposed by \citet{chen2020hybridqa}, is a two-phase pipeline framework to retrieve gold table cells as evidence and input their values and linked passages into a QA model to extract the final answer. \citet{Sun2021IterativeHA} proposes Dochopper, an end-to-end multi-hop retrieval model that directly concatenates rows with related textual evidence as its inputs. \citet{Pan2020UnsupervisedMQ} explores an unsupervised multi-hop QA model, called MQA-QG, which can generate human-like multi-hop questions by building a reasoning graph from heterogeneous data resources. \citet{Kumar2021MultiInstanceTF} propose MITQA, which applies multiple-instance training objectives to retrieve coarse-grained evidence. On the contrary, \citet{Eisenschlos2021MATEMA} introduce a transformer-based model with row- and column-wise attentions for fine-grained evidence retrieval, e.g., table cells. \citet{Wang2022MuGER2ME} propose a unified retriever that tries to preserve the advantages and eliminates the disadvantages of different-granularity evidence retrieval methods. 

TACR differs from the above models mainly in two aspects: (1) TACR focuses on providing an explicit reasoning process by aligning multi-hop questions to tables, so it learns which parts of multi-hop questions are accounted for by retrieving evidence from which modality; and (2) The table-question alignment can enhance the reasoning ability of the table cell selection module with the help of our generated hybrid alignment dataset. TACR shows competitive performance to that of other table QA models on the HybridQA and WTQ datasets on the basis of high row, column, and cell selection accuracy. To the best of our knowledge, no text-table QA system handles the challenge of explicitly showing its reasoning process and multi-hop question table alignment.

\subsection{Table Cell Retrieval}
\citet{jauhar-etal-2016-tables} construct a multiple-choice table QA benchmark that includes over 9000 question-table pairs via crowd-sourcing and proposed a table-cell search model based on calculating all relevance scores between each cell and question. Such a model is reasonable and intuitive but time-consuming. TACR selects gold cells based on row and column selection. Suppose that a table contains $n$ rows and $m$ columns; the table cell search method must calculate $n*m$ scores for each cell, while TACR needs to calculates only $n+m$ scores for each row and column, and selects the gold cell in the row and column with the highest score. \citet{Sun2016TableCS} focus on extracting entities from questions and building a row graph and then mapping the question to the pair of cells in the same row of a table. However, some entities may not appear in both questions and table cells, e.g., an entity of the question in Figure \ref{Figure Example} that should be extracted is \emph{National Football League}, but it cannot be mapped into any cells.

\section{Framework}
\begin{figure*}[ht]
\centering
\includegraphics[width=1.0\textwidth]{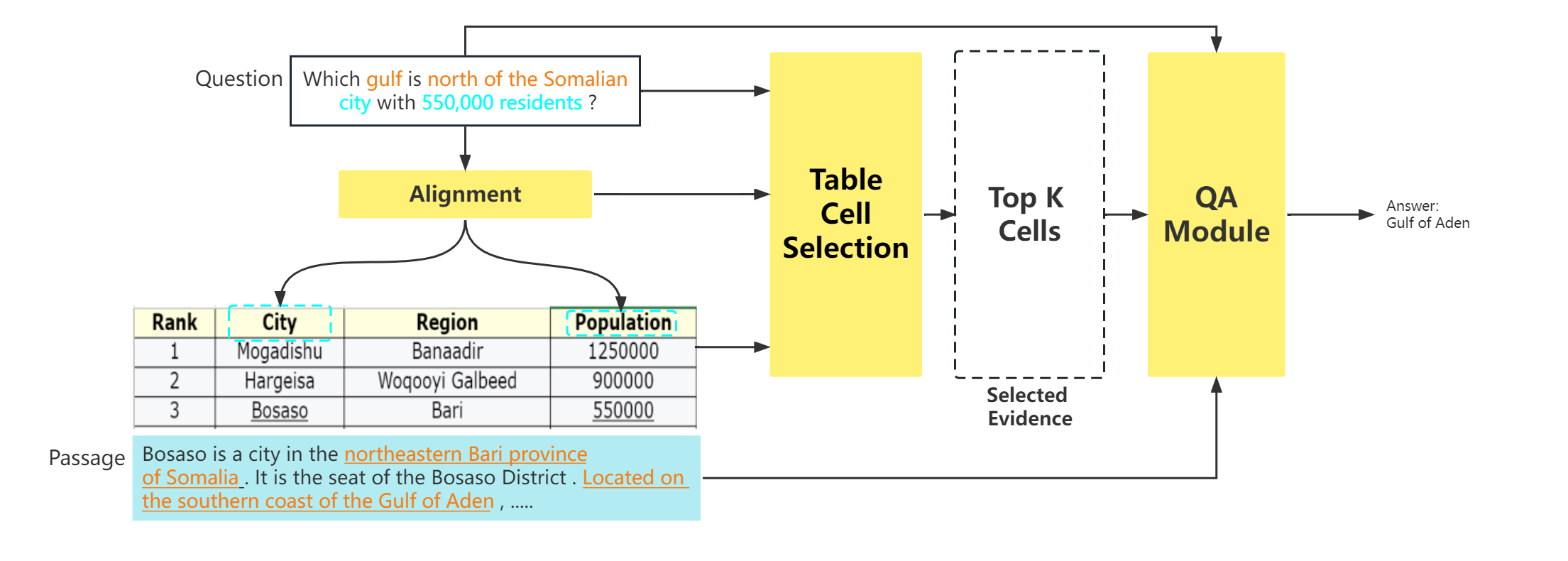}
\caption{TACR model architecture. From left to right, we first construct a hybrid alignment dataset to jointly train the table-question-alignment and table-cell-selection modules. We then concatenate filtered linked passages with selected top-k candidate cells as paragraphs and feed them into a text-based multi-hop QA module to retrieve answers.}
\centering
\label{Figure pipeline}
\end{figure*}

As described in the previous section, both coarse- and fine-grained approaches fail to provide a reasoning process showing which parts of multi-hop questions map to which modality and evidence. Here we describe the details of TACR and its three main components: (1) data augmentation for training the table-question alignment module; (2) a multi-task learning module for table-question alignment and table-cell-selection; and (3) a text-based multi-hop QA module for retrieving answers. Figure \ref{Figure pipeline} shows the overall architecture of TACR.

\subsection{Task Definition}
Given a question $Q$ (a sequence of tokens) and $N$ rows of table $T$ together with linked passages $P$, where each table column has a header $h_{i=1}^{i=M}$ ($M$ is the number of table headers), %. Given the input, 
the task is to find a candidate cell $c_{i,j}$ that contains the answer $\alpha$.

\subsection{Data Construction}
We generate multi-hop questions from tables and linked passages, as well as table-question alignment labels from questions and table columns for training the table-question-alignment module. However, such supervision information is not offered in the HybridQA dataset and other text-table QA datasets, which makes the alignment task difficult. We use an unsupervised text-table QA-generation method to generate questions as well as alignment labels.

\paragraph{Alignment Generation.} We follow the settings of the MQA-QG method~\cite{Pan2020UnsupervisedMQ}, using a pre-trained Google T5~\cite{Raffel2019ExploringTL}, fine-tuned on the SQuAD dataset~\cite{Rajpurkar2018KnowWY}, to generate multi-hop questions from tables and passages based on a bridge entity, a table cell that contains the bridge entity, and a linked passage that describes the bridge entity. The bridge entity is critical in reasoning because it connects the tables and passages, which are difficult to locate in the original HybridQA dataset.

Such bridge entity provides us with additional information to align table headers with generated questions based on the column containing golden cells and the column containing the bridge entity. We align the columns which contain bridge entities and answers to questions following two schema-linking alignment rules:\\

\textbf{Name-based Linking.} This rule refers to exact or partial occurrences of the column/table names in the question, such as the occurrences of “player” in the question in Figure \ref{Figure Example}. Textual matches are the most explicit evidence of table-question alignment and, as such, one might expect them to be directly beneficial to the table-question alignment module. 

\textbf{Value-based Linking.} Table-question alignment also occurs when the question mentions any values that occur in the table and consequently participate in the table-cell selection, such as “the second most” in Figure \ref{Figure Example}. While it is common for examples to make the alignment explicit by mentioning the column name (e.g., “Rank”), many real-world questions do not (like in the example). Consequently, linking a value mentioned in the question to the corresponding column also requires background knowledge.

\begin{figure}[ht]
\centering
\includegraphics[width=0.5\textwidth]{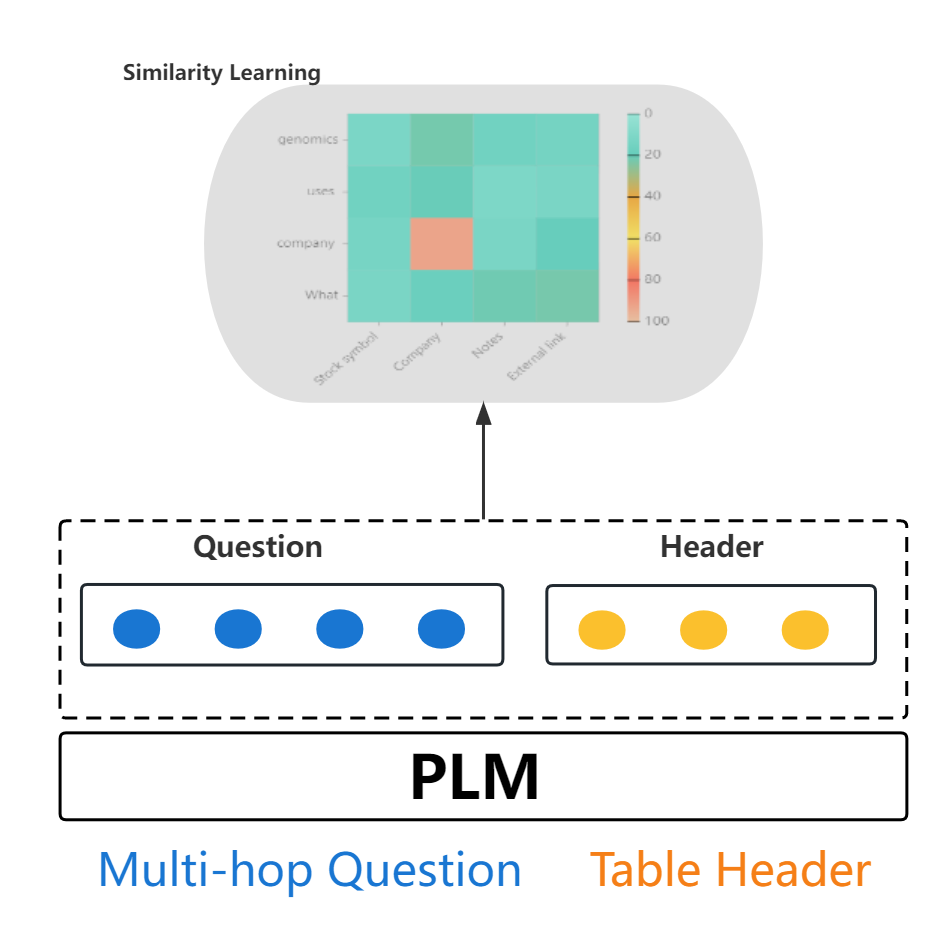}
\caption{The table-question-alignment module of TACR. We treat the alignment objective as a similarity learning task.}
\centering
\label{Figure alignment}
\end{figure}

\subsection{Passage Filtering}
In this stage, we aim to filter out linked passages unrelated to a question, namely keeping almost noise-free passages for the following modules. Moreover, the total number of tokens in passages linked to table cells can be large, exceeding the maximum input sequence length of current LMs. Thus, we utilize Sentence-BERT \cite{Reimers2019SentenceBERTSE} to obtain question and passage embeddings and rank the top-k sentences based on their text similarities. We expand the cells with the filtered top k-related sentences to both fit in the max input length of language models and to preserve the useful information from passages. More details on this stage are provided in Appendix \ref{sec:passage filtration}.

\subsection{Table Alignment \& Cell Selection}
In this stage, we jointly train a multi-task model with the objectives of selecting the expanded cell that contains the answer and table-question alignment to different modalities to enhance the previous objective. TACR accepts the full table as inputs and outputs the probabilities of selected cells based on the probabilities of row and column selection.

\subsubsection{Table-Question Alignment}
Given a natural language question $ Q = \left \{ q_{1},....q_{\left | Q \right |} \right \} $, a table consisting of several column headers $C = \left \{c_{1}....c_{\left | C \right |}\right \} $, and the corresponding table-question alignment labels $L = \left \{ l_{1}, ... l_{\left | C \right |} \right \} $ where $l_{i} \in [0,1]$ (0 meaning the column header is unrelated to the question $Q$ and 1 meaning the column header is related to $Q$). The goal of our table-question alignment module is to learn the relevance between table-column headers and questions. Table-question relations aid TACR by aligning column references in the question to the corresponding table columns.

We first feed the questions and table columns into the pre-trained model and map them into hidden representations. The question and table-column headers can be denoted as $ \left \{ q_{1},....q_{\left | Q \right |} \right \}$ and $ \left \{c_{1}....c_{\left | C \right |}\right \}$, respectively. Our goal is to induce a function $f(q_{i},c_{j})$ to capture the relevance of a question word $q_{i}$ has on the representation of column header $c_{j}$. Figure \ref{Figure alignment} shows the structure of the alignment module.

Specifically, we use ALBERT~\cite{Lan2019ALBERTAL} as the encoder to learn the representations of tables and column headers. Here we concatenate column headers as a pseudo sentence. The representations of the question ({$h_{q}$}) and the column headers sequence ($h_{c}$) are first computed independently. The relevance where each column header $c_{i}$ is the target of the question is then given by using softmax. The respective equations are as follows:
\begin{align}
h_{q} & = \texttt{BERT(Q)}, \\
h_{c} & = \texttt{BERT(C)}, \\
p(C_{i}\in C) & = \texttt{softmax}(W (h_{q} * h_{c}) + b).
\end{align}

\subsubsection{Table-Cell Selection}
Inspired by the previous idea of modeling the attention on rows and columns \cite{Eisenschlos2021MATEMA}, we design a cell-selection module based on row and column selection. The probabilities of each row and column are given and the cells with the top-k highest scores are returned as the candidate answers, or to aid in locating the relevant passage. However, unlike in MATE \cite{Eisenschlos2021MATEMA}, we can derive probabilities of candidate cells from the probabilities of row and column. 

We utilize the Row-Column-Intersection (RCI) model, designed for the single-hop table-QA task~\cite{Glass2021CapturingRA} (based on ALBERT~\cite{Lan2019ALBERTAL}), as our backbone and decompose the table QA task into two subtasks: projection - corresponding to identifying columns; and selection - identifying rows. Every row and column identification is a binary sequence pair classification. We concatenate the question as the first sequence and the row or column as the second sequence. We feed concatenated two sequences, with standard separator tokens $[CLS]$ and $[SEP]$, as the input to the model. The representation of the final hidden state is sent to the linear layer, followed by a softmax to classify whether the column or row contains the answer or not. Each row and column is assigned a probability of containing the answer. This module finally outputs the top-k cells with the sum of row and column probabilities. Therefore, given a table $T$ with $N$ rows and $M$ columns, we can obtain two sets of scores produced from the RCI model: $P_{r} = {p_{1},....p_{N}}$ for rows and $P_{c} = {p_{1},....p_{M}}$ for columns. We then calculate the overall probability score for each cell.

The final training loss is the summation of table-question-alignment loss, table-row-selection loss, and table-column-selection loss:
\begin{align}
    \begin{split}
         L &= \texttt{L\_{row}} + \texttt{L\_{column}} \\
         &+ \sigma \times \texttt{BCE}(pred\_{headers}, target\_{headers}),
    \end{split}
\end{align}
where $\sigma$ is a hyper-parameter to balance cell-selection training and table-question-alignment training. The details of choosing the best $\sigma$ are provided in Appendix \ref{sec:implementation details}.

\subsection{Passage Question-Answering}
Previous research often simply treat the answer-reasoning task as a span-extraction task, considered the first span matching the answer text as the gold span, and use that for training. Such consideration is incorrect because the answer text may appear in multiple passages, but only one of them is right. Therefore, using all text matches for training span extraction may introduce a large amount of training noise. As not all instances are the gold answer text that has relations with questions, after obtaining the top-k cells from the cell-selection module, we train the text-based QA module to predict the final answer that also takes into account the cell-selection scores. 

Specifically, we select clean training instances where the gold answer text appears only once and train an initial QA model. In this stage, we use RoBERTa~\cite{Liu2019RoBERTaAR} as our backbone model. Other BERT variants, e.g., either SpanBERT~\cite{Joshi2019SpanBERTIP} or DeBERTa~\cite{He2020DeBERTaDB}, could be also used in this module. Our goal is to obtain a span $s$ in a given expanded table cell $c$ with its filtered passage $p$ and the input question $q$. We compute a span representation as follows:
%     \begin{center}
       \begin{align}
        & h_{start} = \texttt{RoBERTa}_r(q, c)[\texttt{START}(s)], \\
        & h_{end}  = \texttt{RoBERTa}_r(q, c)[\texttt{END}(s)], \\
        & S_{span}(q, p)  = \texttt{MLP}([h\_{start}, h\_{end}]).
        \end{align} 
%    \end{center}
We also consider other cells in the same row as the retrieved candidate gold cells as the necessary context. We linearize and concatenate the row into a passage with the designed template: "The <column header> is <cell content>". We retrieve the top-k cells and thus have k samples. Since not all selected cells contain the gold answer text, we treat one sample as positive and the others as negative samples. For each data point, we generate $k$ samples %as described in the previous paragraph, 
and match these with the answer text. Let $K = \{ q_{i}, A_{i}, P^{+}_{i}, P^{-}_{i,1}, · · ·, P^{-}_{i,k-1} \}^{k}_{i=1}$ be the training data that consist of $k$ instances, where $k$ is the number of selected candidate cells. Each instance contains one question $q_{i}$, the gold answer text $A_{i}$, and one correct (positive) passage text $P^{+}_{i}$, along with $k-1$ wrong passages $P^{-}_{i,j}$. For positive samples, the answer is the text span of the passage, while for negative samples, the answers are -1. 

\section{Experiments}
\subsection{Datasets}

\textbf{HybridQA}~\cite{chen2020hybridqa} is the first large-scale multi-hop QA dataset that requires reasoning over hybrid knowledge, including tables and linked Wikipedia passages. The dataset contains 62,682 instances in the training set, 3,466 instances in the development set, and 3,463 instances in the test set. 

\textbf{WikiTableQuestions}~\cite{pasupat-liang-2015-compositional}, WTQ for short, consists of 22033 complex questions and 2108 semi-structured Wikipedia tables. The questions are designed by crowd-sourcing to contain a wide range of domains. The answers are derived from several operations such as table lookup, aggregation, superlatives, arithmetic operations, joins, and unions. 

\begin{table}[t]
%\begin{tabular}{lllll}
\begin{tabular}{lrrrr}
\toprule
Split & Train & Dev. & Test & Total \\
\midrule
In-Passage & 35215 & 2025 & 2045 & 39285 \\
In-Table & 26803 & 1349 & 1346 & 29498 \\
Compute & 664 & 92 & 72 & 864 \\
Total & 62682 & 3466 & 3463 & 69611 \\
\bottomrule
\bottomrule
\end{tabular}
\caption{Statistics of HybridQA dataset}
\label{Statistics}
\end{table}

To verify the performance of TACR, we first conduct experiments on HybridQA \cite{chen2020hybridqa}, a dataset of multi-hop question-answering over tabular and textual data. The basic statistics of HybridQA are listed in Table \ref{Statistics}. The dataset contains three partitions: ‘In-Table’, where the answer derives from table cell values; ‘In-Passage’, where the answer exists in a linked passage; and ‘Compute’, where the answer should be computed by executing numerical operations. We mainly focus on the first two types. We also provide results over WTQ to illustrate TACR’s capabilities in table-focused QA.

\subsection{Baselines}
\textbf{MQA-QG}, proposed by \cite{Pan2020UnsupervisedMQ}, is an unsupervised question-generation framework that generates multi-hop questions from tables and linked passages, and uses the generated questions to train an HQA model. \\
\textbf{Table-Only} \cite{chen2020hybridqa} only retrieves the tabular information to find an answer by parsing the question into a symbolic form and executing it. \\
\textbf{Passage-Only} \cite{chen2020hybridqa} only retrieves answers from the table-linked passages. \\
\textbf{Hybrider} \cite{chen2020hybridqa} addresses HQA using a two-stage pipeline framework to retrieve the gold table cell and extract an answer in its value or linked passages.\\
\textbf{Dochopper} \cite{Sun2021IterativeHA} first converts a table with its hyperlinked passages into a long document then concatenates column headers, cell text, and linked passages in each row of tables as a paragraph. \\
\textbf{MATE} \cite{Eisenschlos2021MATEMA} applies sparse attention to rows and columns in a table. To apply it to the HybridQA dataset, the authors propose a PointR module, which expands a cell using the description of its entities, selects the golden cells, then retrieves answers from them. \\
\textbf{MITQA} \cite{Kumar2021MultiInstanceTF} designs a multi-instance training method based on distant supervision to filter the noisy information from multiple answer spans. 

\subsection{Quantitative Analysis}
We use exact match (EM) and F1 scores as evaluation metrics on the HybridQA dataset to compare the performance of TACR with that of previous baselines. As shown in Table \ref{tab:results}, TACR outperforms most baselines and achieved competitive performance to state-of-the-art (SOTA) models (e.g., MITQA) in both EM and F1 scores over the HybridQA dataset. Table \ref{tab:wtq_results} reports the accuracy performance on WTQ. Though TACR is trained on a base model, it presents comparable accuracy to the large SOTA models and outperforms other base models. It is important to note that, besides both using much larger LMs than TACR (GPT-3 and BART-large respectively, versus RoBERTa-base), neither Binder nor Omnitab-large provide explainability. With the help of the table-question-alignment module, TACR boosts relative accuracy by $+18.5\%$ on the test set compared with RCI~\cite{Glass2021CapturingRA}, which is also based on cell selection. This competitive performance is mainly based on the high cell selection along with table-question alignment. We further verified the effectiveness of the table-question-alignment module in an ablation study discussed in Section \ref{sec_ablation}.

\subsection{Qualitative Analysis}
We compare the cell-selection accuracy of TACR and baseline models, as shown in Table \ref{cell retrieval}. The high cell selection accuracy is based on the high row- and column-selection accuracies shown in Table \ref{row column and cell select accuracy}. On the HybirdQA dataset, TACR shows SOTA performance and 0.4\% higher than that of MATE~\cite{Eisenschlos2021MATEMA} in the top 3 cell-selection accuracies due to its 89.3\% row-selection accuracy and 98.3\% column-selection accuracy, as shown in Table \ref{row column and cell select accuracy}. Moreover, by achieving soft question decomposition (i.e., showing which parts of questions are connected to reasoning in the different modalities), TACR both improves the explainability of its results and provides valuable signals for future improvements. 

\begin{table*}[!ht]
\small
\centering
\resizebox{1.0\textwidth}{!}{
\begin{tabular}{l|cccccc|cccccc}
\toprule
\toprule
\multicolumn{1}{l|}{\multirow{1}{*}{Model}} & \multicolumn{6}{c|}{Dev.} & \multicolumn{6}{c}{Test} \\ 
\multicolumn{1}{l|}{} & \multicolumn{2}{c}{In-Table} & \multicolumn{2}{c}{In-Passage} & \multicolumn{2}{c|}{Total} & \multicolumn{2}{c}{In-Table} & \multicolumn{2}{c}{In-Passage} & \multicolumn{2}{c}{Total} \\
& EM & F1 & EM & F1 & EM & F1 & EM & F1 & EM & F1 & EM & F1   \\ 
\midrule
Table-Only & 14.7 & 19.1 & 2.4 & 4.5 & 8.4 & 12.1 & 14.2& 18.8 & 2.6 & 4.7 & 8.3 & 11.7 \\
Passage-Only & 9.2 & 13.5 & 26.1 & 32.4 & 19.5 & 25.1 & 8.9 & 13.8 & 25.5 & 32.0 & 19.1 & 25.0 \\
Hybrider ($\tau$=0.8) & 54.3 & 61.4 & 39.1 & 45.7 & 44.0 & 50.7 & 56.2 & 63.3 & 37.5 & 44.4 & 43.8 & 50.6 \\
\midrule
PointR + SAT & 66.5 & 71.8 & 60.3 & 69.2 & 61.2 & 68.7 & 64.6 & 70.1 & 59.6 & 68.5 & 60.1 & 67.4 \\ %
PointR + TAPAS & 68.1 & 73.9 & 62.9 & 72.0 & 63.3 & 70.8  & 67.8 & 73.2 & 62.0 & 70.9 & 62.7 & 70.0\\  %
\midrule
PointR + TABLEETC & 36.0 & 42.4 & 37.8 & 45.3 & 36.1 & 42.9 & 35.8 & 40.7 & 38.8 & 45.7 & 36.6 & 42.6 \\
PointR + LINFORMER & 65.5 & 71.1 & 59.4 & 69.0 & 60.8 & 68.4 & 66.1 & 71.7   & 58.9 & 67.8 & 60.2 & 67.6 \\ %
PointR + MATE & 68.6 & 74.2 & 62.8 & 71.9 & 63.4 & 71.0 & 66.9 & 72.3 & 62.8 & 71.9 & 62.8 & 70.2 \\ %
\midrule
MQA-QG (unsupervised) & -- & -- & -- & -- & -- & -- & 36.2 & 40.6& 19.8 & 25.0 & 25.7 & 30.5 \\ %
Dochopper & -- & --& -- & -- & 47.7 & 55.0 & -- & -- & -- & --& 46.3 & 53.3 \\ %
MITQA & 68.1 & 73.3 & 66.7 & 75.6 & 65.5 & 72.7 & 68.5 & 74.4 & 64.3 & 73.3 & 64.3 & 71.9 \\ %
MuGER$^2$  & 58.2 & 66.1 & 52.9 & 64.6 &  53.7 & 63.6 & 56.7 & 64.0 & 52.3 & 63.9 & 52.8 & 62.5 \\ %
\midrule
TACR (ours) & 66.7 & 70.3 & 63.4 & 72.5 & 64.5 & 71.6 & 64.1 & 69.6 & 65.4 & 70.7 & 66.2 & 70.2 \\ %
\midrule
Human & & & & & & & & & & & 88.2 & 93.5 \\
\bottomrule
\bottomrule
\end{tabular}}
\caption{EM and F1 results of models on the HybridQA dataset. In-Table and In-Passage subsets refer to the location of answers.}
\label{tab:results}
\end{table*}

\begin{table}[!ht]
\small
\centering
\resizebox{0.5\textwidth}{!}{
\begin{tabular}{l|c|c}
\toprule
\multicolumn{1}{l|}{\multirow{1}{*}{Model}} & \multicolumn{1}{c|}{Dev} & \multicolumn{1}{c}{Test} \\ 
& Acc & Acc \\ 
\midrule	
TAPEX-Large \cite{Liu2021TAPEXTP} & 57.0& 57.5 \\ %
Binder \cite{Cheng2022BindingLM} & 65.0 & 64.6 \\ %
OmniTab-Large \cite{Jiang2022OmniTabPW} & 62.5 & 63.3 \\ %
TAPAS\_base (pre-trained on SQA) \cite{Herzig2020TaPasWS} & - & 48.8 \\ %
UnifiedSKG \cite{Xie2022UnifiedSKGUA} & 50.7 & 49.3 \\ %
TaBERT\_base \cite{Yin2020TaBERTPF} & 51.6 & 51.4 \\%
\midrule
RCI \cite{Glass2021CapturingRA}& 45.3 & 41.7 \\ %
TACR\_RoBERTa-base (ours) & 58.9 & 60.2 \\ %
\bottomrule
\end{tabular}
}
\caption{Execution-accuracy results of models on WTQ}
\label{tab:wtq_results}
\end{table}

\begin{table}[t]
\small
\centering
\resizebox{0.5\textwidth}{!}{
\begin{tabular}{llll}
\toprule
Model & Hits@1 & Hits@3 & Hits@5 \\
\midrule
TABLEETC \cite{ainslie-etal-2020-etc} & 51.1 & 72.0 & 78.9 \\
LINFORMER \cite{Wang2020LinformerSW} & 77.1 & 86.5 & 90.0 \\
MATE \cite{Eisenschlos2021MATEMA} & 80.1 & 86.2 & 90.5 \\
\midrule
TACR (ours) & \textbf{83.3} & \textbf{87.8} & \textbf{91.2} \\
\bottomrule
\end{tabular}}
\caption{ Comparison of cell-retrieval results on HybridQA dataset (dev set)}
\label{cell retrieval}
\end{table}

\subsection{Ablation Study} 
\label{sec_ablation}
To evaluate the impact of the table-question-alignment module, we conduct an ablation study, shown in Table \ref{ablation}. We test DeBERTa-base, ALBERT-base, and RoBERTa-base models as TACR backbones for generality. Different top-k results show that the alignment module consistently significantly improves results; with the best model based on ALBERT improving cell-selection accuracy by 2.5, 3.9, and 4.3\% in top 1, 3, and 5 cell selection respectively; and mean reciprocal rank (MRR) improving by 3.7\%. The results indicate that the table-question-alignment module has an important role in the table-question-reasoning stage to select the most related cells that support the answer to the question.

\begin{table*}[t]
\small
\centering
\resizebox{0.8\textwidth}{!}{
\begin{tabular}{lllll}
\toprule
Model & MRR & Hits@1 & Hits@3 & Hits@5 \\
\midrule
TACR-DeBERT\_base w/o alignment &78.9 & 74.9 & 79.4 & 83.7 \\
TACR-Roberta\_base w/o alignment & 80.7 & 74.3 & 82.6 & 84.4\\
TACR-ALBERT\_base w/o alignment & 80.1 & 77.1 & 82.8 & 85.4 \\
\midrule
TACR-DeBERTa\_base w/ alignment & 82.4 & 78.3 & 83.4 & 86.2 \\
TACR-RoBERTa\_base w/ alignment & 82.5 & 76.5 & 85.5 & 88.9\\
TACR-ALBERT\_base w/ alignment & \textbf{83.8} & \textbf{79.6} & \textbf{86.7} & \textbf{89.7} \\
\bottomrule
\end{tabular}}
\caption{Ablation study of table-question-alignment module impact. Experiment results of cell-retrieval on HybridDQA (dev set) show the effectiveness of this module in the table-cell-selection stage.}
\label{ablation}
\end{table*}

\begin{table}[t]
\small
\centering 
\resizebox{0.5\textwidth}{!}{
\begin{tabular}{c|c c|c c}
\toprule
\multicolumn{1}{c|}{\multirow{2}{*}{Model}} & \multicolumn{2}{c|}{HybridQA} & \multicolumn{2}{c}{WTQ} \\
 & Row & Col & Row & Col \\
\midrule
\multicolumn{5}{c}{top 1} \\
\midrule
TACR\_DeBERTa\_base & 85.1 & 95.3 & 53.2 & 93.9 \\
TACR\_ALBERT\_base & 86.7 & 96.1 & 56.8 & 94.4 \\
TACR\_RoBERTa\_base & 86.0 & 96.2 & 52.3 & 94.7\\
\midrule
\multicolumn{5}{c}{top 3} \\
\midrule
TACR\_DeBERTa\_base & 86.2 & 96.2 & 57.6 & 94.2 \\
TACR\_ALBERT\_base & 88.3 & 97.1 & 62.4 & 95.1 \\
TACR\_RoBERTa\_base & 87.9 & 97.3 & 59.3 & 94.9\\
\midrule
\multicolumn{5}{c}{top 5} \\
\midrule
TACR\_DeBERTa\_base & 87.5 & 97.8 & 59.1 & 94.8 \\
TACR\_ALBERT\_base & 89.9 & 98.3 & 68.1 & 95.4 \\
TACR\_RoBERTa\_base & 89.3 & 98.4 & 64.5 & 95.2 \\
\bottomrule 
\end{tabular}}
\caption{Performance of TACR with different backbone models. Top-k rows and columns selection accuracies on HybridQA and WTQ datasets, where k=1, 3, 5. Results demonstrate the effectiveness of TACR.}
\label{row column and cell select accuracy}
\end{table}

\subsection{Case Study}
To illustrate TACR can successfully learn which parts of tables contain golden cells and which parts of questions are required for reasoning in the different modalities, we choose two examples from the HybridQA development set. Appendix \ref{sec:alignment analysis} includes Figures \ref{Figure case} and \ref{Figure case2} showing their word relevances heatmap and analysis.

The question in Case 1 is \textit{"Who is the athlete in a city located on the Mississippi River ?"}. The concatenated table headers string for the corresponding table is \textit{"Year Score Athlete Place"}. The table-question-alignment module helps TACR learn that header terms \textit{"Athlete"} and \textit{"Place"} have higher relevance to the question than the headers of other columns, thus guiding cell-selection. Figure \ref{Figure case} shows its relevance heatmap. TACR again learns which parts of the question account for retrieving evidence in tables. 

The question in Case 2 is \textit{"What is the middle name of the player with the second most National Football League career rushing yards ?"}. The concatenated table headers string for it is \textit{"Rank Player Team(s) by season Carries Yards Average"}. The table-question-alignment module helps TACR learn that the sub-question \textit{"the player with the second most National Football League career rushing yards"} has a higher relevance to the table headers than that of other parts of the original question, thus guiding modality relevance. Figure \ref{Figure case2} shows its relevance heatmap.

\subsection{Error Analysis}
To further analyze TACR, we also calculate statistics for error cases in the model predictions. The error statistics are based on the development set of HybridQA. Through the cell-selection accuracy statistics in Table \ref{cell retrieval}, we find there are 347 tables whose cells are incorrectly selected. 

To better understand the advantages and disadvantages of table-question alignment-based cell selection, we manually sample and examined 20 such error cases (i.e., where TACR does not provide the correct answer in the correct row, column, and cell position). Out of the 20 samples, we find that five error cases (25\%) are due to requiring numerical reasoning operations that cross several cells (which is out of scope for TACR). The majority of errors, 13 of the remaining incorrect cases, are in the same column with a correct answer while in the wrong row. Only one case is from a different row but the same column with the correct answer and only one incorrect case is in a completely different row and column to the correct answer.

\section{Conclusion}
This paper presents TACR, a \textbf{T}able question \textbf{A}lignment-based \textbf{c}ell selection and \textbf{R}easoning model for hybrid text and table QA, evaluated on the HybridQA and WikiTableQuestions datasets. When answering questions given retrieved table cells and passages, TACR attempts to align multi-hop questions to different modalities for correct evidence retrieval. To enhance the QA module with better table cell-selection and table-question-alignment ability, we construct a hybrid alignment dataset generated from the HybridQA dataset. TACR shows state-of-the-art performance in retrieving intermediate gold table cells and competitive performance on the HybridQA and WikiTableQuestions datasets, while improving output explainability.

\section{Limitations}
In this paper, we focus on the hybrid QA task, where the answers to most questions can be extracted from cell values in tables and linked passages using a reading comprehension model. Although TACR performs well in cell selection, one of its limitations is that it lacks numerical reasoning ability across different cells, such as counting and comparing. To enable TACR to answer numerical questions, we will further develop its numerical reasoning capabilities in future work. Another limitation of TACR is that it shows a strong ability in column selection while performing relatively worse in row selection. For future work, we plan to try to improve its row-selection accuracy.

\bibliography{anthology,custom}

\begin{thebibliography}{41}
\expandafter\ifx\csname natexlab\endcsname\relax\def\natexlab#1{#1}\fi

\bibitem[{Ainslie et~al.(2020)Ainslie, Ontanon, Alberti, Cvicek, Fisher, Pham,
  Ravula, Sanghai, Wang, and Yang}]{ainslie-etal-2020-etc}
Joshua Ainslie, Santiago Ontanon, Chris Alberti, Vaclav Cvicek, Zachary Fisher,
  Philip Pham, Anirudh Ravula, Sumit Sanghai, Qifan Wang, and Li~Yang. 2020.
\newblock \href {https://doi.org/10.18653/v1/2020.emnlp-main.19} {{ETC}:
  Encoding long and structured inputs in transformers}.
\newblock In \emph{Proceedings of the 2020 Conference on Empirical Methods in
  Natural Language Processing (EMNLP)}, pages 268--284, Online. Association for
  Computational Linguistics.

\bibitem[{Chen et~al.(2017)Chen, Fisch, Weston, and
  Bordes}]{chen-etal-2017-reading}
Danqi Chen, Adam Fisch, Jason Weston, and Antoine Bordes. 2017.
\newblock \href {https://doi.org/10.18653/v1/P17-1171} {Reading {W}ikipedia to
  answer open-domain questions}.
\newblock In \emph{Proceedings of the 55th Annual Meeting of the Association
  for Computational Linguistics (Volume 1: Long Papers)}, pages 1870--1879,
  Vancouver, Canada. Association for Computational Linguistics.

\bibitem[{Chen et~al.(2020{\natexlab{a}})Chen, Chang, Schlinger, Wang, and
  Cohen}]{Chen2020OpenQA}
Wenhu Chen, Ming-Wei Chang, Eva Schlinger, William~Yang Wang, and William~W.
  Cohen. 2020{\natexlab{a}}.
\newblock \href {https://doi.org/10.48550/arXiv.2010.10439} {Open question
  answering over tables and text}.
\newblock \emph{ArXiv}, abs/2010.10439.

\bibitem[{Chen et~al.(2019)Chen, Wang, Chen, Zhang, Wang, Li, Zhou, and
  Wang}]{chen2019tabfact}
Wenhu Chen, Hongmin Wang, Jianshu Chen, Yunkai Zhang, Hong Wang, Shiyang Li,
  Xiyou Zhou, and William~Yang Wang. 2019.
\newblock \href {https://doi.org/10.48550/arXiv.1909.02164} {Tabfact: A
  large-scale dataset for table-based fact verification}.
\newblock \emph{arXiv preprint arXiv:1909.02164}.

\bibitem[{Chen et~al.(2020{\natexlab{b}})Chen, Zha, Chen, Xiong, Wang, and
  Wang}]{chen2020hybridqa}
Wenhu Chen, Hanwen Zha, Zhiyu Chen, Wenhan Xiong, Hong Wang, and William Wang.
  2020{\natexlab{b}}.
\newblock {HybridQA}: A dataset of multi-hop question answering over tabular
  and textual data.
\newblock \emph{arXiv preprint arXiv:2004.07347}.

\bibitem[{Cheng et~al.(2022)Cheng, Xie, Shi, Li, Nadkarni, Hu, Xiong, Radev,
  Ostendorf, Zettlemoyer, Smith, and Yu}]{Cheng2022BindingLM}
Zhoujun Cheng, Tianbao Xie, Peng Shi, Chengzu Li, R.K. Nadkarni, Yushi Hu,
  Caiming Xiong, Dragomir~R. Radev, Marilyn Ostendorf, Luke Zettlemoyer,
  Noah~A. Smith, and Tao Yu. 2022.
\newblock \href {https://doi.org/10.48550/arXiv.2210.02875} {Binding language
  models in symbolic languages}.
\newblock \emph{ArXiv}, abs/2210.02875.

\bibitem[{Eisenschlos et~al.(2021)Eisenschlos, Gor, M{\"u}ller, and
  Cohen}]{Eisenschlos2021MATEMA}
Julian~Martin Eisenschlos, Maharshi Gor, Thomas M{\"u}ller, and William~W.
  Cohen. 2021.
\newblock \href {https://doi.org/10.48550/arXiv.2109.04312} {{MATE}: Multi-view
  attention for table transformer efficiency}.
\newblock In \emph{Conference on Empirical Methods in Natural Language
  Processing}.

\bibitem[{Glass et~al.(2021)Glass, Canim, Gliozzo, Chemmengath, Chakravarti,
  Sil, Pan, Bharadwaj, and Fauceglia}]{Glass2021CapturingRA}
Michael~R. Glass, Mustafa Canim, A.~Gliozzo, Saneem~A. Chemmengath, Rishav
  Chakravarti, Avirup Sil, Feifei Pan, Samarth Bharadwaj, and Nicolas~Rodolfo
  Fauceglia. 2021.
\newblock \href {https://doi.org/10.48550/arXiv.2104.08303} {Capturing row and
  column semantics in transformer based question answering over tables}.
\newblock In \emph{North American Chapter of the Association for Computational
  Linguistics}.

\bibitem[{He et~al.(2020)He, Liu, Gao, and Chen}]{He2020DeBERTaDB}
Pengcheng He, Xiaodong Liu, Jianfeng Gao, and Weizhu Chen. 2020.
\newblock \href {https://doi.org/10.48550/arXiv.2006.03654} {{DeBERTa}:
  Decoding-enhanced {BERT} with disentangled attention}.
\newblock \emph{ArXiv}, abs/2006.03654.

\bibitem[{Herzig et~al.(2020)Herzig, Nowak, M{\"u}ller, Piccinno, and
  Eisenschlos}]{Herzig2020TaPasWS}
Jonathan Herzig, Pawel~Krzysztof Nowak, Thomas M{\"u}ller, Francesco Piccinno,
  and Julian~Martin Eisenschlos. 2020.
\newblock \href {https://doi.org/10.18653/v1/2020.acl-main.398} {{TaPas}:
  Weakly supervised table parsing via pre-training}.
\newblock \emph{ArXiv}, abs/2004.02349.

\bibitem[{Jauhar et~al.(2016)Jauhar, Turney, and
  Hovy}]{jauhar-etal-2016-tables}
Sujay~Kumar Jauhar, Peter Turney, and Eduard Hovy. 2016.
\newblock \href {https://doi.org/10.18653/v1/P16-1045} {Tables as
  semi-structured knowledge for question answering}.
\newblock In \emph{Proceedings of the 54th Annual Meeting of the Association
  for Computational Linguistics (Volume 1: Long Papers)}, pages 474--483,
  Berlin, Germany. Association for Computational Linguistics.

\bibitem[{Jiang et~al.(2022)Jiang, Mao, He, Neubig, and
  Chen}]{Jiang2022OmniTabPW}
Zhengbao Jiang, Yi~Mao, Pengcheng He, Graham Neubig, and Weizhu Chen. 2022.
\newblock \href {https://doi.org/10.48550/arXiv.2207.03637} {Omnitab:
  Pretraining with natural and synthetic data for few-shot table-based question
  answering}.
\newblock In \emph{NAACL}.

\bibitem[{Joshi et~al.(2019)Joshi, Chen, Liu, Weld, Zettlemoyer, and
  Levy}]{Joshi2019SpanBERTIP}
Mandar Joshi, Danqi Chen, Yinhan Liu, Daniel~S. Weld, Luke Zettlemoyer, and
  Omer Levy. 2019.
\newblock \href {https://doi.org/10.48550/arXiv.1907.10529} {{SpanBERT}:
  Improving pre-training by representing and predicting spans}.
\newblock \emph{Transactions of the Association for Computational Linguistics},
  8:64--77.

\bibitem[{Joshi et~al.(2017)Joshi, Choi, Weld, and
  Zettlemoyer}]{Joshi2017TriviaQAAL}
Mandar Joshi, Eunsol Choi, Daniel~S. Weld, and Luke Zettlemoyer. 2017.
\newblock \href {https://doi.org/10.48550/arXiv.1705.03551} {Triviaqa: A large
  scale distantly supervised challenge dataset for reading comprehension}.
\newblock In \emph{Annual Meeting of the Association for Computational
  Linguistics}.

\bibitem[{Kumar et~al.(2021)Kumar, Chemmengath, Gupta, Sen, Bharadwaj, and
  Chakrabarti}]{Kumar2021MultiInstanceTF}
Vishwajeet Kumar, Saneem~A. Chemmengath, Yash Gupta, Jaydeep Sen, Samarth
  Bharadwaj, and Soumen Chakrabarti. 2021.
\newblock \href {https://doi.org/10.48550/arXiv.2112.07337} {Multi-instance
  training for question answering across table and linked text}.
\newblock \emph{ArXiv}, abs/2112.07337.

\bibitem[{Lan et~al.(2019)Lan, Chen, Goodman, Gimpel, Sharma, and
  Soricut}]{Lan2019ALBERTAL}
Zhenzhong Lan, Mingda Chen, Sebastian Goodman, Kevin Gimpel, Piyush Sharma, and
  Radu Soricut. 2019.
\newblock \href {https://doi.org/10.48550/arXiv.1909.11942} {{ALBERT}: A lite
  {BERT} for self-supervised learning of language representations}.
\newblock \emph{ArXiv}, abs/1909.11942.

\bibitem[{Liu et~al.(2022)Liu, Hu, Lin, and Wen}]{Liu2022SemanticET}
Aiwei Liu, Xuming Hu, Li~Lin, and Lijie Wen. 2022.
\newblock \href {https://doi.org/10.48550/arXiv.2208.03903} {Semantic enhanced
  text-to-sql parsing via iteratively learning schema linking graph}.
\newblock \emph{Proceedings of the 28th ACM SIGKDD Conference on Knowledge
  Discovery and Data Mining}.

\bibitem[{Liu et~al.(2021)Liu, Chen, Guo, Lin, and Lou}]{Liu2021TAPEXTP}
Qian Liu, Bei Chen, Jiaqi Guo, Zeqi Lin, and Jian-Guang Lou. 2021.
\newblock \href {https://doi.org/10.48550/arXiv.2107.07653} {{TAPEX}: Table
  pre-training via learning a neural {SQL} executor}.
\newblock \emph{ArXiv}, abs/2107.07653.

\bibitem[{Liu et~al.(2019)Liu, Ott, Goyal, Du, Joshi, Chen, Levy, Lewis,
  Zettlemoyer, and Stoyanov}]{Liu2019RoBERTaAR}
Yinhan Liu, Myle Ott, Naman Goyal, Jingfei Du, Mandar Joshi, Danqi Chen, Omer
  Levy, Mike Lewis, Luke Zettlemoyer, and Veselin Stoyanov. 2019.
\newblock \href {https://doi.org/10.48550/arXiv.1907.11692} {{RoBERTa}: A
  robustly optimized {BERT} pretraining approach}.
\newblock \emph{ArXiv}, abs/1907.11692.

\bibitem[{Pan et~al.(2020)Pan, Chen, Xiong, Kan, and
  Wang}]{Pan2020UnsupervisedMQ}
Liangming Pan, Wenhu Chen, Wenhan Xiong, Min-Yen Kan, and William~Yang Wang.
  2020.
\newblock \href {https://doi.org/10.48550/arXiv.2010.12623} {Unsupervised
  multi-hop question answering by question generation}.
\newblock In \emph{North American Chapter of the Association for Computational
  Linguistics}.

\bibitem[{Pasupat and
  Liang(2015{\natexlab{a}})}]{pasupat-liang-2015-compositional}
Panupong Pasupat and Percy Liang. 2015{\natexlab{a}}.
\newblock \href {https://doi.org/10.3115/v1/P15-1142} {Compositional semantic
  parsing on semi-structured tables}.
\newblock In \emph{Proceedings of the 53rd Annual Meeting of the Association
  for Computational Linguistics and the 7th International Joint Conference on
  Natural Language Processing (Volume 1: Long Papers)}, pages 1470--1480,
  Beijing, China. Association for Computational Linguistics.

\bibitem[{Pasupat and Liang(2015{\natexlab{b}})}]{Pasupat2015CompositionalSP}
Panupong Pasupat and Percy Liang. 2015{\natexlab{b}}.
\newblock Compositional semantic parsing on semi-structured tables.
\newblock In \emph{Annual Meeting of the Association for Computational
  Linguistics}.

\bibitem[{Raffel et~al.(2019)Raffel, Shazeer, Roberts, Lee, Narang, Matena,
  Zhou, Li, and Liu}]{Raffel2019ExploringTL}
Colin Raffel, Noam~M. Shazeer, Adam Roberts, Katherine Lee, Sharan Narang,
  Michael Matena, Yanqi Zhou, Wei Li, and Peter~J. Liu. 2019.
\newblock \href {https://doi.org/10.48550/arXiv.1910.10683} {Exploring the
  limits of transfer learning with a unified text-to-text transformer}.
\newblock \emph{ArXiv}, abs/1910.10683.

\bibitem[{Rajpurkar et~al.(2018)Rajpurkar, Jia, and
  Liang}]{Rajpurkar2018KnowWY}
Pranav Rajpurkar, Robin Jia, and Percy Liang. 2018.
\newblock \href {https://doi.org/10.48550/arXiv.1806.03822} {Know what you
  don’t know: Unanswerable questions for squad}.
\newblock In \emph{Annual Meeting of the Association for Computational
  Linguistics}.

\bibitem[{Rajpurkar et~al.(2016)Rajpurkar, Zhang, Lopyrev, and
  Liang}]{rajpurkar2016squad}
Pranav Rajpurkar, Jian Zhang, Konstantin Lopyrev, and Percy Liang. 2016.
\newblock \href {https://doi.org/10.48550/arXiv.1606.05250} {Squad: 100,000+
  questions for machine comprehension of text}.
\newblock \emph{arXiv preprint arXiv:1606.05250}.

\bibitem[{Reimers and Gurevych(2019)}]{Reimers2019SentenceBERTSE}
Nils Reimers and Iryna Gurevych. 2019.
\newblock \href {https://doi.org/10.48550/arXiv.1908.10084} {Sentence-bert:
  Sentence embeddings using siamese bert-networks}.
\newblock \emph{ArXiv}, abs/1908.10084.

\bibitem[{Sun et~al.(2021{\natexlab{a}})Sun, Cohen, and
  Salakhutdinov}]{Sun2021EndtoEndMR}
Haitian Sun, William~W. Cohen, and Ruslan Salakhutdinov. 2021{\natexlab{a}}.
\newblock \href {https://doi.org/10.48550/arXiv.2106.00200} {End-to-end
  multihop retrieval for compositional question answering over long documents}.
\newblock \emph{ArXiv}, abs/2106.00200.

\bibitem[{Sun et~al.(2021{\natexlab{b}})Sun, Cohen, and
  Salakhutdinov}]{Sun2021IterativeHA}
Haitian Sun, William~W. Cohen, and Ruslan Salakhutdinov. 2021{\natexlab{b}}.
\newblock \href {https://doi.org/10.48550/arXiv.2106.00200} {Iterative
  hierarchical attention for answering complex questions over long documents}.

\bibitem[{Sun et~al.(2016)Sun, Ma, He, tau Yih, Su, and Yan}]{Sun2016TableCS}
Huan Sun, Hao Ma, Xiaodong He, Wen tau Yih, Yu~Su, and Xifeng Yan. 2016.
\newblock \href {https://dl.acm.org/doi/10.1145/2872427.2883080} {Table cell
  search for question answering}.
\newblock \emph{Proceedings of the 25th International Conference on World Wide
  Web}.

\bibitem[{Wang et~al.(2019)Wang, Shin, Liu, Polozov, and
  Richardson}]{Wang2019RATSQLRS}
Bailin Wang, Richard Shin, Xiaodong Liu, Oleksandr Polozov, and Matthew
  Richardson. 2019.
\newblock \href {https://doi.org/10.48550/arXiv.1911.04942} {Rat-sql:
  Relation-aware schema encoding and linking for text-to-sql parsers}.
\newblock In \emph{Annual Meeting of the Association for Computational
  Linguistics}.

\bibitem[{Wang et~al.(2020)Wang, Li, Khabsa, Fang, and
  Ma}]{Wang2020LinformerSW}
Sinong Wang, Belinda~Z. Li, Madian Khabsa, Han Fang, and Hao Ma. 2020.
\newblock \href {https://doi.org/10.48550/arXiv.2006.04768} {Linformer:
  Self-attention with linear complexity}.
\newblock \emph{ArXiv}, abs/2006.04768.

\bibitem[{Wang et~al.(2022)Wang, Bao, Duan, Wu, He, and
  Zhao}]{Wang2022MuGER2ME}
Yingyao Wang, Junwei Bao, Chaoqun Duan, Youzheng Wu, Xiaodong He, and Tiejun
  Zhao. 2022.
\newblock \href {https://doi.org/10.48550/arXiv.2210.10350} {{MuGER$^2$}:
  Multi-granularity evidence retrieval and reasoning for hybrid question
  answering}.
\newblock \emph{ArXiv}, abs/2210.10350.

\bibitem[{Wolf et~al.(2020)Wolf, Debut, Sanh, Chaumond, Delangue, Moi, Cistac,
  Rault, Louf, Funtowicz, Davison, Shleifer, von Platen, Ma, Jernite, Plu, Xu,
  Le~Scao, Gugger, Drame, Lhoest, and Rush}]{wolf-etal-2020-transformers}
Thomas Wolf, Lysandre Debut, Victor Sanh, Julien Chaumond, Clement Delangue,
  Anthony Moi, Pierric Cistac, Tim Rault, Remi Louf, Morgan Funtowicz, Joe
  Davison, Sam Shleifer, Patrick von Platen, Clara Ma, Yacine Jernite, Julien
  Plu, Canwen Xu, Teven Le~Scao, Sylvain Gugger, Mariama Drame, Quentin Lhoest,
  and Alexander Rush. 2020.
\newblock \href {https://doi.org/10.18653/v1/2020.emnlp-demos.6} {Transformers:
  State-of-the-art natural language processing}.
\newblock In \emph{Proceedings of the 2020 Conference on Empirical Methods in
  Natural Language Processing: System Demonstrations}, pages 38--45, Online.
  Association for Computational Linguistics.

\bibitem[{Xie et~al.(2022)Xie, Wu, Shi, Zhong, Scholak, Yasunaga, Wu, Zhong,
  Yin, Wang, Zhong, Wang, Li, Boyle, Ni, Yao, Radev, Xiong, Kong, Zhang, Smith,
  Zettlemoyer, and Yu}]{Xie2022UnifiedSKGUA}
Tianbao Xie, Chen~Henry Wu, Peng Shi, Ruiqi Zhong, Torsten Scholak, Michihiro
  Yasunaga, Chien-Sheng Wu, Ming Zhong, Pengcheng Yin, Sida~I. Wang, Victor
  Zhong, Bailin Wang, Chengzu Li, Connor Boyle, Ansong Ni, Ziyu Yao,
  Dragomir~R. Radev, Caiming Xiong, Lingpeng Kong, Rui Zhang, Noah~A. Smith,
  Luke Zettlemoyer, and Tao Yu. 2022.
\newblock \href {https://doi.org/10.48550/arXiv.2201.05966} {{UnifiedSKG}:
  Unifying and multi-tasking structured knowledge grounding with text-to-text
  language models}.
\newblock \emph{ArXiv}, abs/2201.05966.

\bibitem[{Yang et~al.(2018)Yang, Qi, Zhang, Bengio, Cohen, Salakhutdinov, and
  Manning}]{yang-etal-2018-hotpotqa}
Zhilin Yang, Peng Qi, Saizheng Zhang, Yoshua Bengio, William Cohen, Ruslan
  Salakhutdinov, and Christopher~D. Manning. 2018.
\newblock \href {https://doi.org/10.18653/v1/D18-1259} {{H}otpot{QA}: A dataset
  for diverse, explainable multi-hop question answering}.
\newblock In \emph{Proceedings of the 2018 Conference on Empirical Methods in
  Natural Language Processing}, pages 2369--2380, Brussels, Belgium.
  Association for Computational Linguistics.

\bibitem[{Yin et~al.(2020)Yin, Neubig, tau Yih, and Riedel}]{Yin2020TaBERTPF}
Pengcheng Yin, Graham Neubig, Wen tau Yih, and Sebastian Riedel. 2020.
\newblock \href {https://doi.org/10.48550/arXiv.2005.08314} {{TaBERT}:
  Pretraining for joint understanding of textual and tabular data}.
\newblock \emph{ArXiv}, abs/2005.08314.

\bibitem[{Yu et~al.(2018)Yu, Zhang, Yang, Yasunaga, Wang, Li, Ma, Li, Yao,
  Roman, Zhang, and Radev}]{Yu2018SpiderAL}
Tao Yu, Rui Zhang, Kai-Chou Yang, Michihiro Yasunaga, Dongxu Wang, Zifan Li,
  James Ma, Irene~Z Li, Qingning Yao, Shanelle Roman, Zilin Zhang, and
  Dragomir~R. Radev. 2018.
\newblock \href {https://doi.org/10.48550/arXiv.1809.08887} {Spider: A
  large-scale human-labeled dataset for complex and cross-domain semantic
  parsing and text-to-sql task}.
\newblock In \emph{Conference on Empirical Methods in Natural Language
  Processing}.

\bibitem[{Zhao and Yang(2022)}]{Zhao2022TablebasedFV}
Guangzhen Zhao and Peng Yang. 2022.
\newblock \href {https://doi.org/10.48550/arXiv.2204.08753} {Table-based fact
  verification with self-labeled keypoint alignment}.
\newblock In \emph{International Conference on Computational Linguistics}.

\bibitem[{Zhong et~al.(2017)Zhong, Xiong, and Socher}]{zhong2017seq2sql}
Victor Zhong, Caiming Xiong, and Richard Socher. 2017.
\newblock \href {https://doi.org/10.48550/arXiv.1709.00103} {Seq2sql:
  Generating structured queries from natural language using reinforcement
  learning}.
\newblock \emph{arXiv preprint arXiv:1709.00103}.

\bibitem[{Zhong et~al.(2018)Zhong, Xiong, and Socher}]{Zhong2018Seq2SQLGS}
Victor Zhong, Caiming Xiong, and Richard Socher. 2018.
\newblock \href {https://doi.org/10.48550/arXiv.1709.00103} {Seq2sql:
  Generating structured queries from natural language using reinforcement
  learning}.
\newblock \emph{ArXiv}, abs/1709.00103.

\bibitem[{Zhu et~al.(2021)Zhu, Lei, Huang, Wang, Zhang, Lv, Feng, and
  Chua}]{zhu2021tat}
Fengbin Zhu, Wenqiang Lei, Youcheng Huang, Chao Wang, Shuo Zhang, Jiancheng Lv,
  Fuli Feng, and Tat-Seng Chua. 2021.
\newblock \href {https://doi.org/10.48550/arXiv.2105.07624} {{TAT-QA}: A
  question answering benchmark on a hybrid of tabular and textual content in
  finance}.
\newblock \emph{arXiv preprint arXiv:2105.07624}.

\end{thebibliography}
\bibliographystyle{acl_natbib}

\appendix
\label{sec:appendix}
\section{Passage Filtering}
\label{sec:passage filtration}
Passage filtering plays an important role in cell selection as well as answer extraction. Pre-trained language models such as BERT, RoBERTa, and LLMs have the limitation of max input sequence length. Passage filtering ensures that it is unlikely to lose information relevant to the questions, while fitting model input limits. We used the well-trained DistilBert-based model to obtain question and passage embeddings to rank and filter relevant passages.\footnote{https://huggingface.co/sebastian-hofstaetter/distilbert-dot-tas\_b-b256-msmarco}

\section{Alignment Analysis}
\label{sec:alignment analysis}

\begin{figure*}[ht]
\centering
\includegraphics[width=1.0\textwidth]{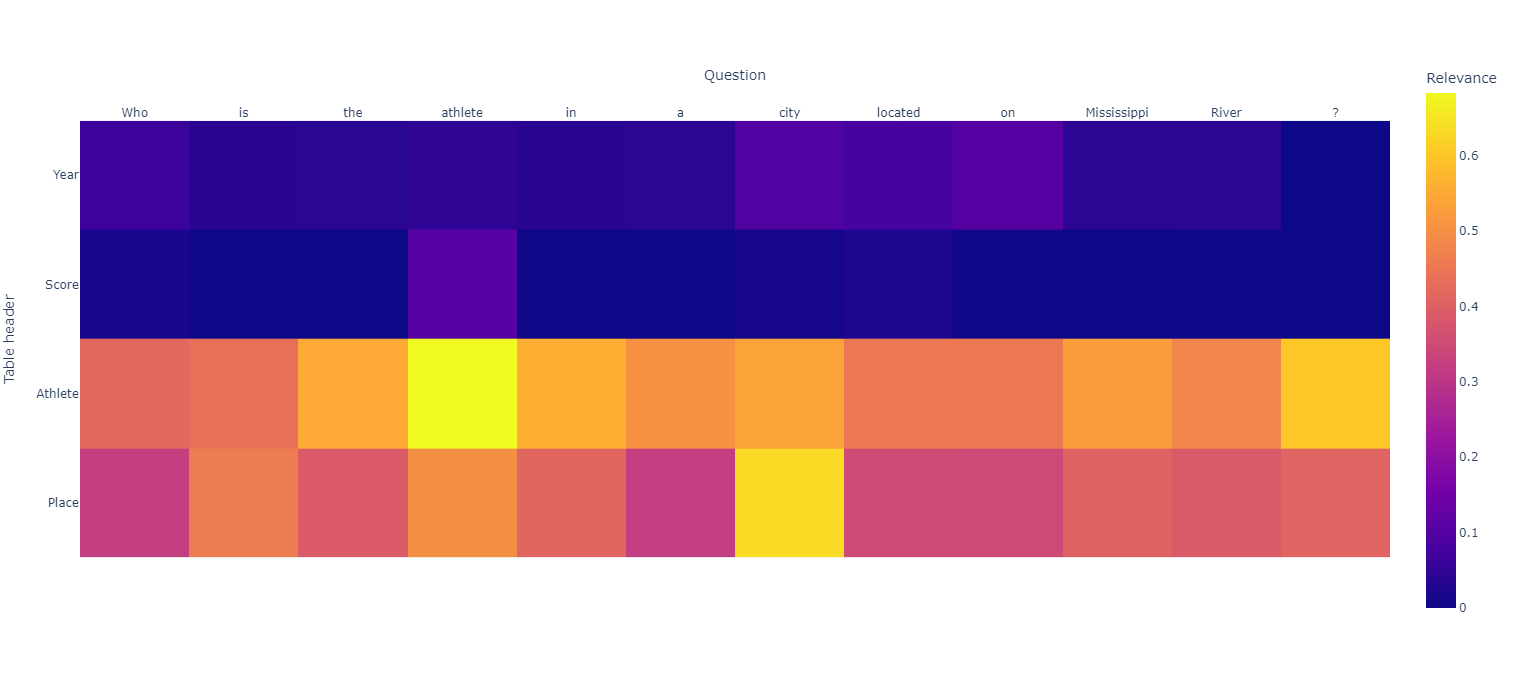}
\caption{Heatmap of question and table-header relevance - Case 1}
\centering
\label{Figure case}
\end{figure*}

\begin{figure*}[t]
\centering
\includegraphics[width=1.0\textwidth]{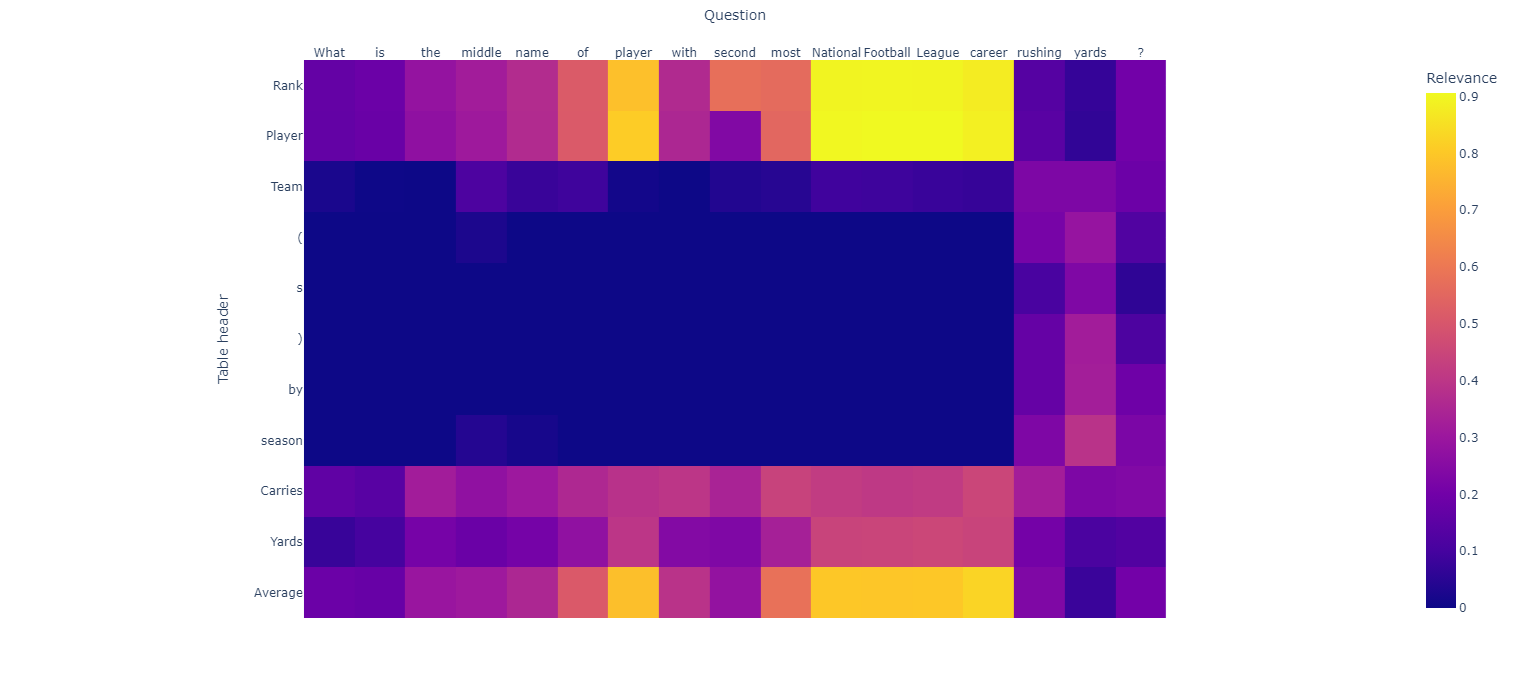}
\caption{Heatmap of question and table header relevance - Case 2}
\centering
\label{Figure case2}
\end{figure*}

Here we provide example heatmaps showing the relevance of questions and table headers. The relevance is in the [0,1] range, where the higher relevance between words from questions and column headers is shown in the warmer colors and vice versa. Figure \ref{Figure case} shows that the column headers "athlete" and "place" have more relevance to the question, which helps TACR identify which columns contain potential gold cells. In Figure \ref{Figure case2}, the words "player with second most national football league" from the question have more relevance to columns, which help TACR learn which parts of the question better use to retrieve gold cells.

\section{Implementation Details of Cell Selection and Alignment} 
\label{sec:implementation details}
TACR is implemented using Pytorch version 1.13 and the Huggingface transformers \cite{wolf-etal-2020-transformers} library. We trained TACR using two NVIDIA A6000 GPUs. The cell selection and table–question-alignment modules are trained for four epochs and we selected the best model based on the dev fold performance. AdamW is used as optimizer algorithm with a learning rate of 5×10-5 and a batch size of 32. We set the per-GPU train batch size to 16 while training the span-based QA model. Final answers are evaluated using EM and F1 scores. We also automatically iterated through increments of 0.1 in the range [0, 1] to select the best $\sigma$ to balance the multi-task training. 

\paragraph{Hyper-parameter Details:} We tune hyper-parameters based on the loss on the development set and use the following range of values for selecting the best hyper-parameters: \\
• Batch size: [8, 16, 32, 64] \\
• Learning rate: [1e-3, 1e-4, 1e-5, 1e-6, 3e-3, 3e-4, 3e-5, 3e-6, 5e-3, 5e-4, 5e-5, 5e-6] \\
• $\sigma$ : [0.1, 0.2, 0.3, 0.4, 0.5, 0.6, 0.7, 0.8, 0.9, 1.0]

\end{document}